# Applying Semantic Segmentation to Autonomous Cars in the Snowy Environment

Zhaoyu PAN*, Takanori EMARU[#], Ankit RAVANKAR[#], Yukinori KOBAYASHI[#]

*Graduate School of Engineering, Hokkaido University, Sapporo, Japan

[#]Research Faculty of Engineering, Hokkaido University, Sapporo, Japan

**Abstract**

This paper mainly focuses on environment perception in snowy situations which forms the backbone of the autonomous driving technology. For the purpose, semantic segmentation is employed to classify the objects while the vehicle is driven autonomously. We train the Fully Convolutional Networks (FCN) on our own dataset and present the experimental results. Finally, the outcomes are analyzed to give a conclusion. It can be concluded that the database still needs to be optimized and the favorable algorithm should be proposed to get the better results.

**Index term:** Fully Convolutional Networks, Semantic Segmentation, database

## 1.  Introduction

Nowadays, both researchers and enterprises emphasize the self-driving field, such as giants Google, Tesla or Chinese Baidu companies. Autonomous technology promises to get more and more mature with the coming of Artificial Intelligence (AI) era, with rapid improvement in deep-learning or Simultaneous Localization and Mapping (SLAM) techniques. Among these techniques, semantic segmentation for autonomous driving technology is extremely significant. Semantic segmentation is a task that classifies the same type of objects into one class [2]. Through the vehicle camera or the laser RADAR, the images are input into the neural network. Then, the car's computer automatically classifies the images to avoid obstacles such as pedestrians and other vehicles. As a result, the car can be driven in the traffic lane correctly and safely. An example of this practical application can be observed in Figure 2.

However, the challenge of this project is how to achieve recognizing all the objects covered by the snow only using the camera data. The difficulty is because the traffic lanes are usually covered by the heavy snow as shown in the left bottom picture in Figure 2. For snowy regions like Hokkaido in Japan, snowy winter can significantly reduce the performance of vision sensors as visibility is diminished for objects to be detected. So far, there are no effective solutions to how the vehicle would adapt to seasonal changes such as heavy snow. The lane and road sign detection would normally fail in such conditions.

This paper focuses on how to apply semantic segmentation to autonomous vehicles in the snowy environment. The special premise forms the backbone of the self-driving technology. In this paper, given that convolutional network is the basic structure among all deep-learning algorithms, Fully Convolutional Networks (FCN) is given priority into consideration [5]. In Section 2, some related works are reviewed including common algorithms of semantic segmentation and databases. After that, the experiments are described and experimental results based on our own created database are shown in Section 3. Sensitivity analyses of the results are also presented through evaluation method. Finally, Section 4 concludes the paper.

## 2.  Related Work

### 2.1 Algorithms of semantic segmentation

Semantic segmentation aims to recognize and classify a categorical label to every pixel in the given image. In the field of semantic segmentation, convolutional networks are driving advances in recognition. Especially, convolutional neural networks (CNNs) have already shown tremendous achievements in the classification area, emerging network structures such as Visual Geometry Group (VGG) and ResNet [3]. Yet, there is one disadvantage in the process of convolution and pooling. Because the feature map will get smaller and some features are lost, it's difficult to do the segmentation precisely. It means that CNNs cannot distinguish the specific contour of the object [4]. Therefore, Fully Convolutional Networks (FCN) came into being and was firstly proposed in 2015. FCN is trained end-to-end, pixels-to-pixels on semantic segmentation [5][6]. FCN can be achieved on the base of mainstream deep convolutional neural networks (CNNs), which means FCN takes advantage of existing CNNs as powerful visual models. One obvious difference between the two algorithms is that FCN takes the place of the fully connected layer into deconvolution layer. The framework of FCN is shown in Figure 1. Above all, FCN becomes the cornerstone of deep-learning applied to semantic segmentation [5].

Based on FCN, Cambridge proposed SegNet. This algorithm aims to solve the issues related to semantic segmentation of self-driving or intelligent robot areas. SegNet consists of encoder network, decoder network and pixel-wise classification layer. One feature of FCN is that FCN utilizes skip structures to obtain more feature details. Comparing to FCN just replicates the encoder characteristics, SegNet duplicates the maximum polling index. SegNet adds the sub-sampling of the max pooling as the decoder. To the extent, it makes SegNet more ef-ficient than FCN in terms of memory usage, in reverse, FCN is much more flexible than SegNet [5][6].

**2.2 Database**

Database always plays the important role in the deep learning. Throughout the years, there are a lot of datasets, such as PASCAL VOC (2012), the Cambridge-driving Labeled Video Database (CamVid) and German Traffic Sign Recognition Benchmark (GTSRB).

**2.2.1 PASCAL VOC (2012)**

The PASCAL VOC (2012) is one of the standard databases for the semantic segmentation. It has 21 classes and its train/validation data contain 11,530 images. Besides, the size of segmentation has substantially increased since 2005 and goes to stable until 2012. FCN network was proposed and tested mainly based on PASCAL VOC. The images in the database are all about the aero plane, bicycle, bird, boat, bottle, bus, car, cat, chair, cow, dining table, dog, horse, motorbike, person, potted plant, sheep, sofa, train, tv/monitor [2]. However, semantic segmentation of the self-driving requires the database of the traffic pictures. And the pictures should be taken during driving the car. It means PASCAL VOC is not suitable for this project.

**2.2.2 GTSRB**

GTSRB is the large and lifelike database which contains more than 50, 000 Germany traffic sign images in 43 classes. This database is created to solve the single-image, multi-class classification problem and only contains physical traffic sign instances. All the images within the database are stored in PPM format and the sizes vary between $15 \times 15$ to $250 \times 250$ pixels. To achieve the self-driving, the road scene images included in the database are also essential, not only the traffic sign images. In another word, this database can't completely support self-driving research.

**2.2.3 CamVid**

The CamVid is the first collection of videos with road information. The dataset contains around 367 images for training, 101 images for validation, 233 images for testing and all the images are with *.png* format in 32 classes: void, building, wall, tree, vegetation, fence, sidewalk, parking block, column/pole, traffic cone, bridge, sign, miscellaneous text, traffic light, sky, tunnel, archway, road, road shoulder, lane markings (driving), lane markings (non-driving), animal, pedestrian, child, cart luggage, bicyclist, motorcycle, car, SUV/pickup/truck, truck/bus, train and other moving objects [2]. This database is appropriate for regular autonomous vehicle research, but not relevant to the snowy environment. Because there are the street scenes without snowy circumstance.

**3. Experiments**

**3.1 New Database**

Since there is no specific existing database fitting to autonomous car in snowy environment, the new database is created for this project. In the database, annotations include 20 classes, which are divided up into the categories car, ego vehicle, static object, sky, roadway, sidewalk, snow mass, vegetation, person, animal, building, traffic sign, traffic light, telegraph pole, truck, bus, field, snow blowing and manhole. The images are taken mainly in Sapporo city in winter with HD resolution of $1080 \times 1920$, $738 \times 1280$, $720 \times 1280$ pixels. The training set includes 524 images, the validation set includes 175 images and the test set includes 174 images. An example of these images is pictured in Figure 2.

**3.2 Training**

Training the FCN on new database was done through GeForce GTX 1080Ti, and the total memory is trained on Ubuntu 16.04. The codes are programmed in Keras framework with Python 2.7.

This topic is a totally new field for semantic segmentation. It is uncertain that which autonomous technology is suitable to the snowy environment. Therefore, FCN is applied at the beginning, although SegNet shows more advantages in the present practical applications. The structure of FCN is based on VGG16 according to the paper [5] and is shown as Figure 1.

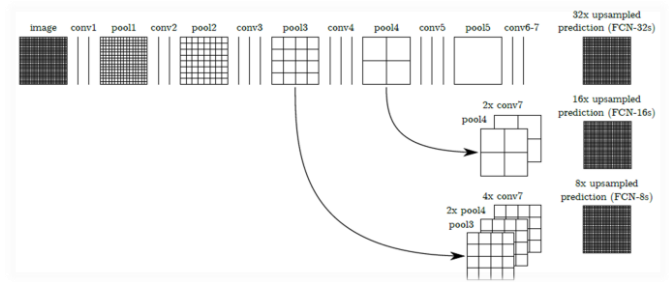

**Fig.1** The architecture of the FCN-8 (Conv = convolutional layer/ pool = pooling layer) [3]

**3.3 Results**

In this experiment, the FCN was trained with

$$Batch\_size = 17, 2, 1$$

$$Epoch = 70, 70, 7$$

Figure 2 shows the example of the semantic segmentation performed on our own test dataset.

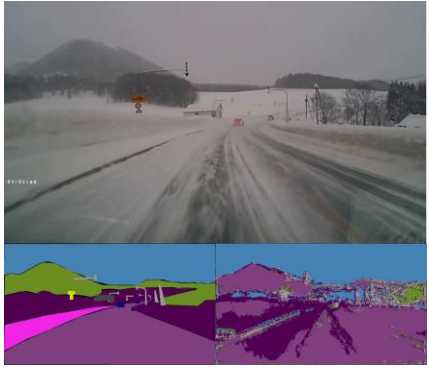

**Fig.2** The top picture shows the original image, one of the training datasets; the picture on the left side is the ground truth, one of the labelled images; the right side is one of the prediction images after the test step.

After the training with different batch sizes and epochs, the results of accuracy and loss are shown as figure 3.

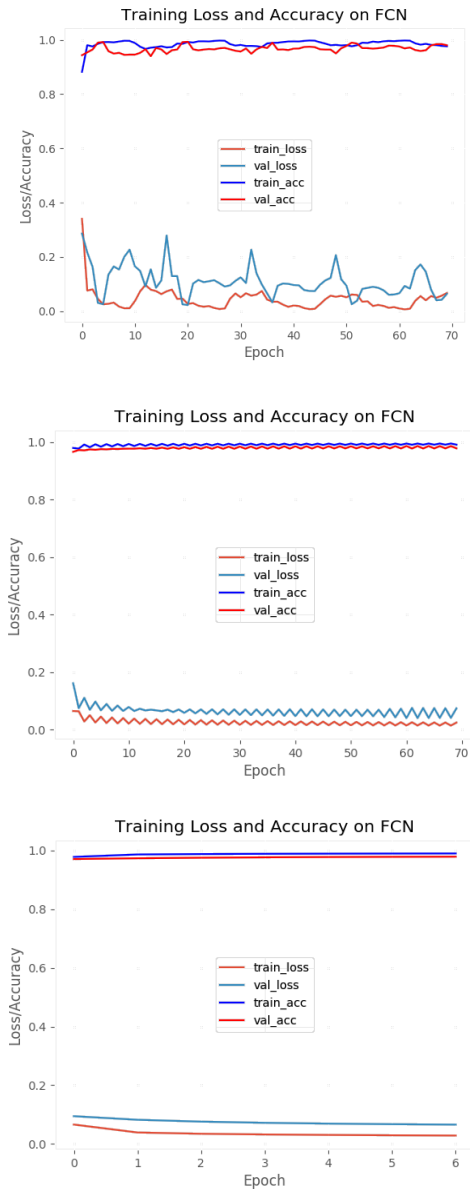

**Fig.3** the pictures show the training results ranged from $Epoch = 70, Batch\_size = 17$,

$Epoch = 70, Batch\_size = 2$, and

$Epoch = 7, Batch\_size = 1$

To evaluate the testing results, this experiment mainly makes use of the quality metric $IoU$ [7]. $IoU$ (Intersection-over-union) is defined as follows:

$$IoU = \frac{DetectionResult \cap GroundTruth}{DetectionResult \cup GroundTruth}$$
$$= \frac{TP}{TP+FP+FN}$$

In which per class $TP$ means true positive, $FP$ stands for false positive and $FN$ represents for false negative. In table 1, the results of testing are displayed relating to each class.

**Table 1.** *Performance $IoU$ of the FCN on new created database with the different training steps. (Nan means not a number)*

|   | Class | BS=17 Epochs=70 | BS=2 Epochs=70 | BS=1 Epochs=7 |
|---|---|---|---|---|
| 1 | Manhole | Nan | Nan | Nan |
| 2 | Unlabeled | 0.0009 | 0.0011 | 0.0005 |
| 3 | Ego-vehicle | 0. | 0. | Nan |
| 4 | Static object | 0.0979 | 0.1090 | 0.0293 |
| 5 | Car | 0.0070 | 0.0589 | 0.0022 |
| 6 | Sky | 0.7273 | 0.7767 | 0.7520 |
| 7 | Roadway | 0.4326 | 0.5528 | 0.4911 |
| 8 | Sidewalk | 0. | Nan | Nan |
| 9 | Snow mass | Nan | Nan | 0 |
| 10 | Vegetation | 0.6606 | 0.6867 | 0.6567 |
| 11 | Person | Nan | 0. | 0 |
| 12 | Animal | Nan | Nan | Nan |
| 13 | Building | 0.2763 | 0.2447 | 0.2047 |
| 14 | Traffic sign | 0.0008 | 0.0686 | 0.0034 |
| 15 | Traffic light | 0. | 0. | 0.0061 |
| 16 | Telegraph pole | 0.0048 | 0.0549 | 0.0579 |
| 17 | Truck | 0. | 0.0008 | 0.0001 |
| 18 | Bus | Nan | Nan | 0 |
| 19 | Field | 0. | 0. | Nan |
| 20 | Snow blowing | 0.4280 | 0.5079 | 0.4493 |
| | **Prediction Time (s/per pic)** | **0.1227** | **0.1376** | **0.1225** |

### 3.4 Discussion

From the first picture of Figure 3, at the beginning of training, both validation accuracy and training accuracy are increasing with the training epoch, in contrast, the training and validation loss are decreasing. And it can also be concluded from the third picture. However, there is an obvious overfitting with the continuous increasing of epoch. When $Epoch = 70, Batch\_size = 2$, the overfitting phenomenon is becoming apparent from the second picture of Figure 3.

Due to that it's better to have less prediction time for autonomous vehicle, the prediction time is calculated and recorded in the Table 1. The average of prediction time is

$$T_{pred} = 0.1276 \text{ s/pic},$$

and it can be acceptable.

The value of $IoU$ in Table 1 shows that sky has the highest accuracy rate. Roadway, vegetation and snow blowing three classes are also having the higher accuracy rate than other classes. Moreover, there are some classes which don't have the $IoU$ value. One reason of $IoU$ value showing nan is because the datasets don't include this class, such as animal. In winter environment, it's rare to see the animals on the road in Hokkaido.

The training results prove that there exists the most suitable training epoch, otherwise it will cause the overfitting in the process of training. Finally, it may also affect the final accuracy of prediction. Although the class roadway presents the higher accuracy rate, the value of $IoU$ is around 0.50, which is still much lower than the requirement which can meet the self-driving technique.

## 4. Conclusion and Future Work

The results of this work show that FCN based on our own dataset leads to a relatively low detection rate. The reasons caused to such results are the database and the neural network itself two aspects. For the database, it has its own limitation, for example, most pictures are taken in the similar environment. It means for the neural network these pictures are almost the same picture in one extent. Moreover, the database doesn't contain every defined class. For this class, such as animal, the accuracy would be nan, more than zero. On the other hand, FCN is not a good enough algorithm to be applied to do the semantic segmentation in the snowy environment so far. Combing with other algorithms will be instrumental and effective, such as CRF. Or trying other neural network structures may be also helpful for self-driving in the snowy environment.

In one word, this paper explains an application example, which is that FCN is trained to do semantic segmentation for autonomous vehicles in the snowy environment.

## Acknowledgement

This work was supported by "Project to support the upgrading of strategic core technology", The Hokkaido Bureau of Economy, Trade, and Industry, Japan.

## REFERENCES

[1] Martin Thoma, "A Survey of Semantic Segmentation". Computer Vision and Pattern Recognition, arXiv:1602.06541 or arXiv:1602.06541v2, 2016

[2] Alberto Garcia-Garcia, Sergio Orts-Escolano, Sergiu Oprea, Victor Villena-Martinez, Jose Garcia-Rodriguez, "A Review on Deep Learning Techniques Applied to Semantic Segmentation". Computer Vision and Pattern Recognition, arXiv:1704.06857, 2017

[3] J. Niemeijer, P. Pekezou Fouopi, S. Knake-Langhorst, and E. Barth, "A Review of Neural Network based Semantic Segmentation for Scene Understanding in Context of the self driving Car". Student Conference on Medical Engineering Science, BioMedTec Studierendentagung, 07.-09. März 2017, Lübeck.

[4] Bike Chen, Chen Gong, Jian Yang, and Member, IEEE, "Importance-Aware Semantic Segmentation for Autonomous Vehicles". IEEE Intelligent Transportation Systems Society, 10.1109/TITS.2018.2801309.

[5] Jonathan Long, Evan Shelhamer, Trevor Darrell, "Fully Convolutional Networks for Semantic Segmentation". Computer Vision and Pattern Recognition, arXiv:1605.06211, 2016.

[6] Vijay Badrinarayanan, Alex Kendall, Roberto Cipolla, "SegNet: A Deep Convolutional Encoder-Decoder Architecture for Image SegmentationS". IEEE Transactions on Pattern, Volume: 39 Issue: 12, 2016.

[7] Gabriela Csurka, Diane Larlus, Florent Perronnin, "What is a good evaluation measure for semantic segmentation?". CSURKA, LARLUS, PERRONNIN: EVALUATION OF SEMANTIC SEGMENTATION, report number: 2013/027, 2013. Available: http://www.bmva.org/bmvc/2013/Papers/paper0032/paper0032.pdf